\documentclass[sigconf]{acmart}
\AtBeginDocument{%
  }

\usepackage{multirow}
\usepackage{multicol}
\usepackage{booktabs}
\usepackage{adjustbox}

\setcopyright{acmlicensed}
\copyrightyear{2025}
\acmYear{2025}
\acmDOI{XXXXXXX.XXXXXXX}
\acmConference[Conference acronym 'XX]{Make sure to enter the correct
  conference title from your rights confirmation email}{June 03--05,
  2025}{Woodstock, NY}
\acmISBN{978-1-4503-XXXX-X/2018/06}

\begin{document}

\title{Dynamic Rank Adaptation for Vision-Language Models}


\author{Jiahui Wang}
\email{wjh@stu.ahu.edu.cn}
\affiliation{%
  \institution{Anhui University}
  \city{Hefei}
  \country{China}
}

\author{Qin Xu}
\affiliation{%
  \institution{Anhui University}
  \city{Hefei}
  \country{China}}
  \authornotemark[1]
\email{xuqin@ahu.edu.cn}

\author{Bo Jiang}
\affiliation{%
  \institution{Anhui University}
  \city{Hefei}
  \country{China}}
  \authornote {Corresponding Authors}
\email{zeyiabc@163.com}

\author{Bin Luo}
\affiliation{%
  \institution{Anhui University}
  \city{Hefei}
  \country{China}}
\email{luobin@ahu.edu.cn}
\renewcommand{\shortauthors}{Anonymous Authors}

\begin{abstract}
Pre-trained large vision-language models (VLMs) like CLIP demonstrate impressive generalization ability. Existing prompt-based and adapter-based works have made significant progress in fine-tuning VLMs but still face the challenges of maintaining strong generalization abilities, particularly towards unseen new classes. This limitation partly arises from these methods treating all tokens of the image and text encoder equally, which can lead to overfitting on less informative features (e.g., background noise, template words) and degrade the general representations that are crucial for novel concept recognition. To address this issue, we propose Dynamic Rank Adaptation (DRA), a novel adapter variant method, designed specifically to enhance new class generalization. DRA dynamically allocates adaptation ranks based on the importance of features during training to preserve general knowledge. DRA first employs token importance grouping, using sequence attention to evaluate and group tokens by their importance. Then, we adopt rank adaptation according to the importance of each token group dynamically by assigning higher feature ranks to the more important tokens. Also, we design a new channel response mechanism to prioritize the preservation and adaptation of feature channels identified as the most informative for each instance. In addition, a L1 regularization term is introduced to stabilize the training. Extensive experiments demonstrate the effectiveness and superiority of our proposed DRA over existing works, especially on enhancing the performance of new classes on various benchmarks, including base-new classes, cross-datasets evaluation and domain generalization. The source code will be published after the paper is received. 
\end{abstract}

\ccsdesc[500]{Computing methodologies~Computer vision}

\keywords{Vision-Language Models, Dynamic Rank, Model Adaptation, Transfer Learning.}



\maketitle

\section{Introduction}
Pre-trained large vision-language models (VLMs) \cite{clip,align,blip,filip}, serving as foundational models, have recently shown significant progress. Typically trained under supervision on large-scale image-text pair data, the most representative work, CLIP \cite{clip}, employs a contrastive learning strategy to map image information and text descriptions into a shared feature space. During inference, hand-crafted prompts, such as "a photo of a [\texttt{class}]", are often used for tasks like classification through image-text matching. Benefiting from large-scale pre-training, these VLMs generally exhibit excellent generalization capabilities across a range of downstream tasks without the need for fine-tuning \cite{zero1,lit,zero2,zero3}.

Although large VLMs have shown impressive generalization capabilities, the large amount of training data and large model parameter size required hinder their further fine-tuning in downstream tasks. Prompt learning, as a widely used method, is often used to transfer pre-trained VLMs to downstream tasks. The core idea is to introduce task-specific learnable prompt tokens in the text encoder \cite{cocoop,coop, kgcoop, tcp}, image encoder \cite{vpt}, or both \cite{maple,promptsrc}. During training, only prompt tokens are optimized, while other parameters of the model remain frozen. For example, CoOp \cite{coop} proposed replacing hand-designed prompts with learnable prompts and optimizing them in fine-tuning. MaPLe \cite{maple} proposed adding learnable prompts to both text branch and  visual branch and using prompts to build interactions between modalities. ArGue \cite{argue} proposed to generate several attribute prompts for each category to improve the discriminant ability of the model for each category. Prompt learning greatly reduces the computational requirements and provides a practical method to transfer pre-trained VLMs for downstream tasks. 
Beyond prompt learning, another pipeline of work adapts pre-trained large models to downstream tasks by fine-tuning lightweight external modules which are known as "Adapters" \cite{adaptervision,adapternlp, adaptformer, tipadapter,dynaadapt}. 
Adapter architectures like Clip-Adapter \cite{clipadapter} and AdaptFormer \cite{adaptformer} insert Adapter modules at the final layer or intermediate layers of the frozen network, enabling parameter-efficient fine-tuning. LoRA \cite{lora} proposes performing a low-rank decomposition of the weight updates for linear layers and merging the result with the original feature path, utilizing re-parameterized techniques to avoid additional computational overhead during inference. MMA \cite{mma} introduces Adapters into the image and text branches separately and facilitates cross-modal information interaction by sharing an MLP between the two branches. Compared to prompt learning, adapter-based methods often offer greater flexibility when applied to a broader range of backbone networks. 

\begin{figure}
    \centering
    \includegraphics[width=1\linewidth]{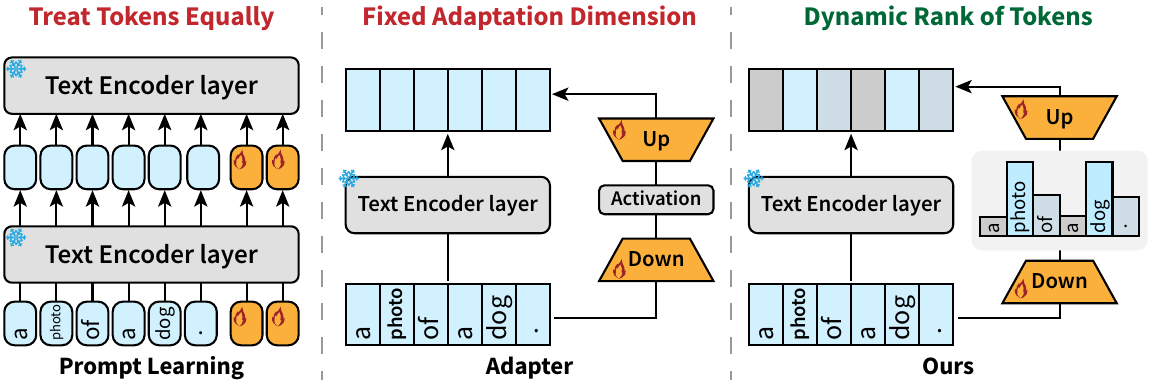}
    \caption{The motivation of the proposed method is to dynamically assign different ranks to different tokens and reduce overfitting of irrelevant information.}
    \label{fig:motivation}
\end{figure}

However, despite the significant progress of these fine-tuning methods, such as Prompt learning and Adapters, they often struggle to maintain strong generalization capabilities, particularly when encountering unseen new classes after being fine-tuned on the base classes \cite{dept}. 
These fine-tuning methods can significantly enhance the model's performance on base classes due to their robust fitting capabilities, but often at the expense of generalization capabilities on unseen new classes. This phenomenon limits the potential of the model for application in open scenarios. We observe that existing fine-tuning methods generally give equal attention to all input tokens during the adaptation process, as shown in Figure \ref{fig:motivation}. 
Excessive adaptation to tokens with low informational value or weak task relevance (such as image background noise or blank word vectors in text sequences) not only introduces computational redundancy but may also compromise the model's generalization capability due to the forgetting of common features that are critical for generalization. 


Based on the above observations, this paper aims to optimize the VLM fine-tuning process by introducing dynamic importance weighting to specifically enhance cross-class generalization capabilities while maintaining foundational model competencies. 
We propose a novel Adapter variant method called Dynamic Rank Adaptation (DRA). The core idea of DRA is to overcome the fixed-rank constraint in adapter design, dynamically allocating adaptation capacity according to the significance/importance of input text and image feature tokens.  
Specifically, we first employ a token attention mechanism to assess the importance of each token. This score indicates the likelihood that each token carries discriminative information within the current sample.  
Based on these scores, we sort the tokens and divide them into different importance groups. 
Then, DRA dynamically
allocates adaptation ranks based on the importance of features
to preserve the general knowledge during training, i.e., it adopts rank adaptation dynamically
by assigning higher feature ranks to the more important tokens. 
In addition, we develop a new channel response mechanism to prioritize the preservation and adaptation of feature channels identified as the most informative for each instance. Finally, a L1 regularization term is introduced to stabilize the whole training. 
We verify the effectiveness of DRA on a wide range of benchmarks, including generalization from base classes to new classes, few-shot learning, and domain adaptation scenarios. Our main contributions can be summarized as follows.
\begin{itemize}
    \item We propose to identify and analyze the challenge of new class generalization in VLM fine-tuning from the perspective of token importance and propose Dynamic Rank Adaptation to enhance model generalization ability. 
    \item We propose the token importance grouping enables the adapter to allocate adaptation capacity based on input feature importance dynamically for each sample.
    \item We propose the channel response mechanism to preferentially preserve useful information when adapting features with lower ranks.
\end{itemize}


\section{Related Works}
\label{sec:related}
\subsection{Vision-Language Models}
Recent years have witnessed remarkable progress in Vision-Language Models (VLMs) that learn joint representations from large-scale image-text data \cite{clip, align, filip, blip}. The well-known CLIP \cite{clip} stands out due to its effective contrastive learning strategy, aligning images and text descriptions in a shared embedding space. CLIP models that are pre-trained on massive datasets \cite{laion5b} exhibit impressive zero-shot generalization capabilities across various downstream tasks, often requiring only hand-crafted text prompts (e.g., "a photo of a [\texttt{class}]") for classification. However, the huge parameters of these models and the vast pre-training data make full fine-tuning computationally prohibitive for many applications. Therefore, efficiently adapting pre-trained VLMs like CLIP to specific downstream tasks, particularly in few-shot settings, has become a critical research area \cite{clipadapter, tipadapter, promptdet, regionclip, opendet, clipseg, extractclip}. Our work builds upon CLIP and aims to improve its adaptation for better generalization.

\subsection{Prompt Learning in VLMs}
Prompt learning is the mainstream approach for the efficient fine-tuning of VLMs. Research in NLP \cite{promptnlp} shows that adding task-sharing prompts to pre-trained models can effectively improve the transferability of downstream tasks. This line of work introduces the handcraft templates, or learnable vectors named prompts, into the input space of the text encoder \cite{coop,cocoop, kgcoop, prograd, tcp}, visual encoder \cite{comma}, or both \cite{maple, promptsrc}, optimizing only these prompts during fine-tuning. CoOp \cite{coop} proposed to use learnable prompts by replacing handcrafted text prompts with continuous learnable vectors, showing strong performance in few-shot scenarios. CoCoOp \cite{cocoop} extended this by making prompts conditional on the input image, aiming for better instance-specific adaptation. KgCoOp \cite{kgcoop} proposed to regularize the text features by learnable prompts with handcraft prompts, while ProGrad \cite{prograd} focuses on gradient alignment to preserve pre-trained knowledge. PLOT \cite{plot} proposed to match local features with the text features for fine-grained semantic alignment by introducing optimal transport distance. MaPLe \cite{maple} introduces prompts in both modalities and links them to enable cross-modal learning. Prompt-based methods significantly improve the performance of VLMs on base classes. However, during fine-tuning on downstream tasks, task-shared prompts cannot be dynamically adapted to each sample and treat all input tokens with the same weight. This may lead to overfitting on base classes and reduce the generalization ability to unknown concepts.

\subsection{Adapter in Transfer Learning}
Another popular strategy in transfer learning involves inserting small, trainable networks named adapter into the frozen layers of the pre-trained model \cite{adaptervision, adapternlp}. Adapters typically consist of lightweight MLP layers with bottleneck structures with lower intermediate dimensions \cite{adaptformer}. ClipAdapter \cite{clipadapter} adds adapters and conducts fine-tuning with the output feature of CLIP on either the image or text branch. AdapterFormer \cite{adaptformer} integrates adapters within the middle layers of the transformer blocks. LoRA \cite{lora} proposes to perform optimization by learning a low-rank representation of the original weight matrix, which can be merged back during the inference process to avoid increasing the inference cost. TIP-adapter \cite{tipadapter} proposes to use the visual features of labeled samples of each category as the initial weights of the adapter to assist model classification. MMA \cite{mma} employs adapters for both image and text branches and shares MLP for modality interaction. However, the fixed feature ranks for all inputs and tokens ignore the different importance of different features (e.g., foreground vs. background, keywords vs. template words). Although DyT \cite{dynaadapt} proposed to skip unimportant tokens during adaptation, simply mask these tokens will cause the performance degradation of base class. Existing fixed rank design is considered a potential bottleneck for generalization, especially for new categories, as it may discard the discriminative information of important tokens or overfit to less informative tokens. We address this issue by introducing dynamic rank adaptation within the adapter framework by dynamically allocate adaptation feature rank and prioritize preserving high response channels of unimportant tokens.
 
\begin{figure*}[ht]
    \centering
    \includegraphics[width=1\linewidth]{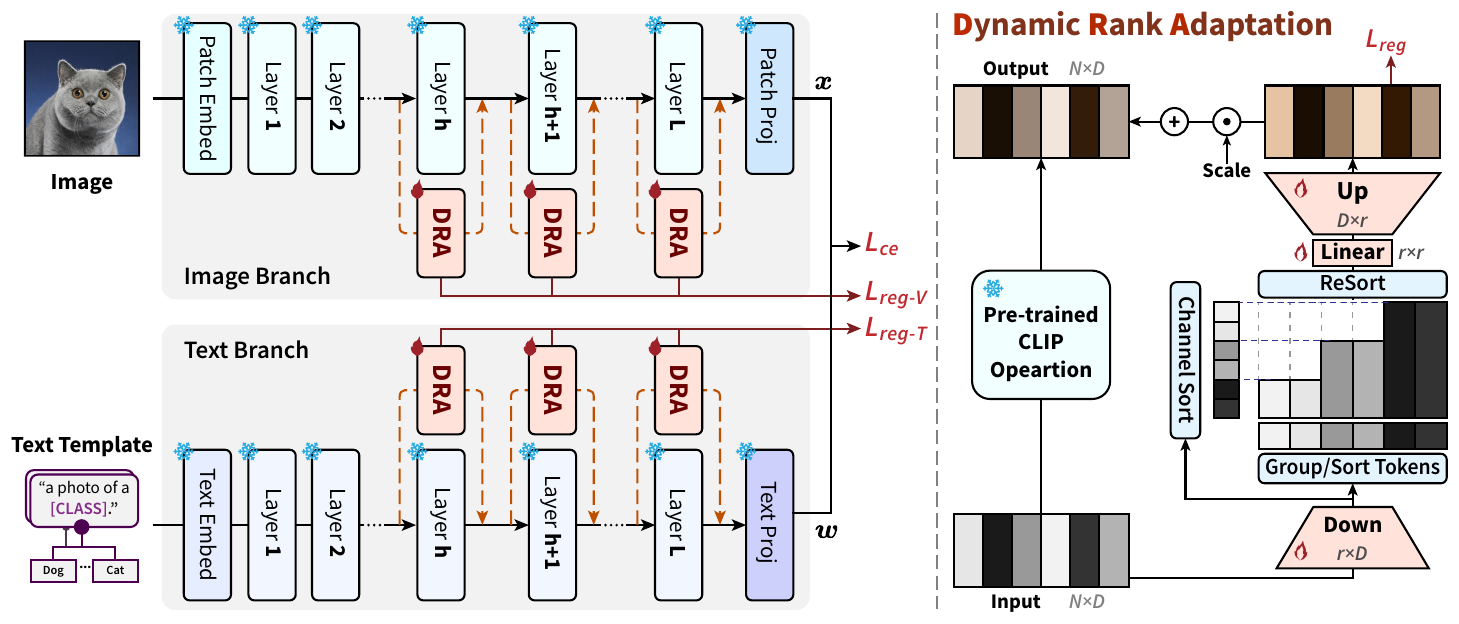}
    \caption{The diagram of the proposed Dynamic Rank Adaptation (Left) and the Detailed view (Right). 
    DRA dynamically assigns different ranks to the adaptation feature based on the importance of the tokens and prioritizes preserving important channels. DRA are applied to the frozen image and text encoders strat from $h$ layer.}
    \label{fig:main}
\end{figure*}

\section{Method}
\label{sec:method}

\subsection{Preliminary}
\label{sec:clip}
\subsubsection{CLIP Model}
Similar to prior work, our research focuses on the few-shot generalization task for CLIP models. In this setting, the model is first trained on base classes from various datasets using limited samples, and subsequently evaluated directly on unseen new classes. CLIP is a foundational vision-language model typically pre-trained on extensive paired image-text samples. Its architecture comprises two main components: an image encoder $V$ and a text encoder $T$. For the image encoder, given an input image $I$, features are derived as follows:
\begin{align}
    X_0 &= \mathrm{Embed}_V(I)\in \mathbb{R}^{M\times d_v}\\
[c_i,X_i] &= V_i([c_{i-1},X_{i-1}])\quad i = 1,2,\dots,L\\
x &= \mathrm{FC}_V(c_L)
\end{align}
where $\mathrm{Embed}_V$ denotes the image embedding layer, $M$ is the number of image tokens, $L$ is the number of layers, and $c_0$ is a trainable classification token concatenated to the input sequence. The projection layer $\mathrm{Proj}_V$ maps the final class token $c_L$ to the image feature $x$. In the text branch, class-specific prompts are generated using templates $\mathcal{T}$ such as "a photo of a $\texttt{[class]}$", where $\texttt{[class]}$ represents the category name. These text prompts are processed as:
\begin{align}
    W_0 &= \mathrm{Embed}_T(\mathcal{T})\in \mathbb{R}^{C\times N\times d_t}\\
W_i &= T_i(W_{i-1})\quad i =  1,2,\dots, L\\
w &= \mathrm{FC}_T(W_L)
\end{align}
where $\mathrm{Tokenize}_T$ converts text to token sequences, $N$ is the token sequence length, $C$ is the number of categories, and $\mathrm{Proj}_T$ generates the final text feature $w$ from the encoder output $W_L$. During inference, similarity scores between the visual feature $x$ and text feature $w_i$ for each class $i$ are computed via cosine similarity to predict class probabilities:
\begin{equation}
    p(y=i|x) = \frac{\exp\left(\cos(x, w_i) / \tau\right)}{\sum_{j=1}^C \exp\left(\cos(x, w_j) / \tau\right)}
\end{equation}

where $\tau$ is a learnable temperature parameter, and $\cos(\cdot, \cdot)$ denotes cosine similarity. $y$ represents the predicted class label.

\subsubsection{Adapter} 
Adapter architectures are widely used for fine-tuning large pre-trained models, typically consisting of a few MLP layers. Specifically, a common Adapter structure, as seen in \cite{adaptformer}, first projects feature to a lower dimension $r$ using linear weights $W_d \in \mathbb{R}^{d \times r}$, and then projects them back to the original dimension $d$ using another linear weights $W_u \in \mathbb{R}^{r \times d}$. The output of the Adapter is usually added to the output of the frozen network module via a residual connection:
\begin{equation}
    X_{\text{adapter}} = \mathrm{OP}(X) + s\cdot\sigma(\sigma(XW_d)W_u)
\end{equation}
where $d$ is the original feature dimension, $r$ is the intermediate dimension ($r \ll d$), $\sigma$ represents the ReLU activation function, $\mathrm{OP}(X)$ is the output of the original frozen network module (e.g., FFN or attention block), and $s$ is a scaling factor. The effectiveness of Adapters is often attributed to the low-rank property of features, i.e. their effective informational dimension is relatively low in over-parameterized deep learning models \cite{lora}. However, existing Adapter-based methods apply the same fixed rank $r$ to all tokens when adapting models to downstream tasks, neglecting the varying information density across different tokens.

\subsection{Dynamic Rank Adaptation}
For the few-shot generalization task with CLIP, the network faces a dual challenge: learning discriminative knowledge for base classes while preserving generalization ability to novel classes. Current methods often treat all tokens in text/image sequences equally, overlooking significant differences in semantic importance, such as between foreground/background regions or keywords/template words. For instance, standard Adapters applying a uniform low-rank bottleneck to all tokens might lead to information loss for crucial tokens or introduce redundant noise from less important tokens. This can cause the overfitting phenomenon, particularly in few-shot scenarios, thereby impairing generalization to novel classes.

To this end, we propose Dynamic Rank Adaptation (DRA), which dynamically allocates higher effective feature ranks of tokens with greater information content while compressing less informative tokens. DRA reduces the overfitting of irrelevant information and enhances the generalization ability of the model to novel classes. For example, in image tokens, foreground regions typically carry denser semantic information and should retain a higher rank, while background regions are applied with more aggressive low-rank compression. In text tokens, class keywords warrant high-rank representations, while template words and padding tokens can be rank-reduced. Furthermore, to alleviate the potential impact of rank reduction on base class performance, our proposed DRA dynamically retains channels that exhibit higher responses to the current sample during dimensionality reduction. The overall diagram of our method is illustrated in Figure \ref{fig:main}.
\subsubsection{Token Importance Grouping}
First, we need to assess the importance of each token within the sequence. Similar to standard adapters, we project the input features $X$ with the rank of $\min(N, d)$ to a low-rank representation $X'$ using the down-projection layer $W_d$:
\begin{equation}
    X' = \sigma(XW_d) \in \mathbb{R}^{N \times r}
\end{equation}
Then, we calculate token importance scores $a_t \in\mathbb{R}^N$ within this intermediate dimension $r$ using an attention mechanism:
\begin{equation}
    a_t = \frac{1}{N}\sum^N_{i=0} \mathrm{Softmax}\left(\frac{(X'W_t)(X'W_t)^\top}{\sqrt{d}} \right)_{ij}\in \mathbb{R}^{N}
\end{equation}
where $W_t \in \mathbb{R}^{r \times r}$ is a learnable projection matrix for encoding token importance information, $N$ is the sequence length, and $r$ is the intermediate feature dimension. By averaging the columns of the attention matrix, we obtain an importance score $a_t$ for each token. Tokens with higher scores are considered received more attention of other tokens and thus more important during fine-tuning. We adopt this score as the basis for token importance grouping by sort the score to obtain a sorted index list $t$:
\begin{equation}
    t = \arg \mathrm{sort}(a_t + g)
\end{equation}
where $g \sim \text{Gumbel}(0, 1)$ is a noise vector of the same dimension sampled from the Gumbel distribution, ensuring the sorting process is differentiable during training.

\subsubsection{Channel Response} Reducing the rank for less important token group might discard useful information, potentially harming base class performance. To mitigate this, we introduce a channel response mechanism that prioritizes retaining channels in the intermediate dimension $r$ that are more active for the current sample. We use the variance of each channel as a measure of its response. A larger variance indicates that the token sequence exhibits a wider spread of values in that channel, suggesting it potentially carries more information and is thus more important. The channel importance score $a_c$ is calculated as follows:
\begin{equation}
    a_c = \frac{1}{N-1}\mathrm{diag}\left((X'W_c-\mu)^\top(X'W_c-\mu)\right) \in \mathbb{R}^{r}
\end{equation}
where $W_c \in \mathbb{R}^{r \times r}$ is a learnable projection matrix, and $\mu$ is the mean of $X'W_c$ along the sequence length dimension. Subsequently, we sort the $r$ channels for each sample based on $a_c$ to obtain a channel index list $c$:
\begin{equation}
    c = \arg \mathrm{sort}(a_c + g)
\end{equation}
Indices appearing earlier in $c$ correspond to channels with higher response and potentially more relevant information for the sample.
\subsubsection{Rank Adaptation}
Based on the token importance order $t$, we partition the tokens uniformly into $K$ groups. Groups containing tokens with higher importance scores are assigned a higher effective rank, while lower-importance groups are assigned a lower dimension. Specifically, let each group contain $k=N/K$ tokens. We group the token indices based on the importance sort order $t$:
\begin{equation}
    G_i = \{t_{[i \cdot k + 1]}, \dots, t_{[i \cdot k + k]} \} \quad i = 0, 1, \dots, K-1
\end{equation}
We define a rank ratio $e_i \in (0, 1]$ for each group $G_i$, where higher importance groups correspond to larger $e_i$ values. For the $i$-th group, according to its rank ratio $e_i$ and the channel response order $c$, we retain the top $\lfloor e_i \times r \rfloor$ most active channels, constructing a group-level channel mask $\mathcal{M} \in \mathbb{R}^{K \times r}$:
\begin{equation}
    \mathcal{M}_{ij} = \begin{cases}
 1/e_i &, j \in c_{[0:\left \lfloor e_i\times r \right \rfloor ]} \\
 0 &, \mathrm{otherwise}
\end{cases}
\quad i\in 1,\dots K, j\in1\dots r
\end{equation}
where $r$ is the intermediate dimension and the maximum available rank. Note that, for training stability, we rescale the retained channels by the inverse of their ratio $e_i$, similar to the scaling operation in Dropout. Next, we expand the group-level masks $\mathcal{M}$ to the full sequence length by the number of tokens in each group to obtain token-level masks $\mathcal{M}' \in \mathbb{R}^{N \times r}$. We then apply this mask to the sorted low-dimensional features $X'$ and conduct resort operation to get the rank-adapted features $X^*$:
\begin{equation}
    X^* = \mathrm{RestoringSort}(\mathrm{Sort}(X', t) \odot \mathcal{M}'), t)
\end{equation}
Finally, we project the rank-adapted features $X^*$ back to the original dimension using the up-projection layer and fuse them with the output of the original module:
\begin{equation}
    \widetilde{X} = \mathrm{OP}(X) + s \cdot \sigma(\sigma(X^* W_p) W_u)
\end{equation}
where $W_p \in \mathbb{R}^{r \times r}$ and $W_u \in \mathbb{R}^{r \times d}$ are learnable projection  matrices, $\sigma$ is the ReLU activation function, and $s$ is the scaling factor. We introduce the additional linear layer $W_p$ to enhance the representational capacity. It allows the model to learn how to better combine and refine the important feature channels that were dynamically selected and retained before projecting them back to the high-dimensional space. Notably, despite the inclusion of $W_p$, the varying ranks of different token groups are preserved, as the ultimate rank is determined by the dynamic adaptation process, i.e., the zero values columns in $\mathcal{M}'$ and $r$. Matrix multiplication and ReLU operation do not increase the rank.

Our proposed DRA is designed to reduce fitting to irrelevant information during training and is enabled only during the training phase. During inference, features are processed directly through linear projections and fused output of the original module,
\begin{equation}
    \overline{X} = \mathrm{OP}(X) + s \cdot \sigma(\sigma(\sigma(XW_d) W_p) W_u)
\end{equation}
The same DRA strategy is employed in both the image and text branches, but they operate independently.

\subsection{Regularization Loss}
To stabilize the learning process of the Adapter, we introduce a regularization term. This term encourages the DRA output during training, $\widetilde{X}$, to keep close to the inference-time output, $\overline{X}$, measured using the L1 normalization:
\begin{equation}
    \mathcal{L}_{reg} = \| \widetilde{X} - \overline{X} \|_1 = \sum_{i=1}^{N} \sum_{j=1}^{d} |\widetilde{X}_{ij} - \overline{X}_{ij}|
\end{equation}
Drawing to the insights from works like \cite{mma,skip}, which suggest that fine-tuning lower layers of CLIP offers limited benefit for base classes and can harm novel class generalization, we insert the DRA modules into the encoder layers starting from the $h$-th layer in both image and text branches, as shown in Figure \ref{fig:main}. We compute $\mathcal{L}_{reg}$ for each Adapter where DRA is applied and sum them up separately for the image and text branches to obtain $\mathcal{L}_{reg-V}$ and $\mathcal{L}_{reg-T}$. The final overall loss function is defined as:
\begin{equation}
    \mathcal{L} = \mathcal{L}_{ce}(p(y|x), y') + \lambda_T \mathcal{L}_{reg-T} + \lambda_V \mathcal{L}_{reg-V}
\end{equation}
where $\mathcal{L}_{ce}$ is the standard cross-entropy loss, $y'$ is the ground-truth label, and $\lambda_V$ and $\lambda_T$ are hyperparameters to balance the loss.

\section{Experiment}
\label{sec:experiment}

\begin{table*}[ht]
    \caption{Comparison Results with SOTA methods 
    in the Base-New generalization setting. “Base” and “New” are the accuracies on base and novel classes respectively. “HM” is the harmonic mean of base and new accuracy.}
    \label{tab:main}
    \centering
    \setlength{\tabcolsep}{3pt}
    \begin{adjustbox}{width=\linewidth}
    \begin{tabular}{rccc|ccc|ccc|ccc|ccc|ccc} 
    \toprule
         \multirow{2}{*}{Method} &  \multicolumn{3}{c|}{Average}&  \multicolumn{3}{c|}{ImageNet}&  \multicolumn{3}{c|}{Caltech101}&  \multicolumn{3}{c|}{OxfordPets} & \multicolumn{3}{c|}{StanfordCars} & \multicolumn{3}{c}{Flowers102}\\ 
          &Base & New & HM& Base & New & HM& Base & New & HM&Base & New & HM & Base & New & HM & Base & New &HM \\
\midrule
         CLIP \cite{clip}&  69.34& 74.22&71.70&  72.43& 68.14&70.22&  96.84& 94.00&95.40&  91.17& 97.26&94.12
 & 63.37& 74.89& 68.65& 72.08& \textbf{77.80}&74.83\\ 
         CoOp \cite{coop}&   82.69& 63.22&71.66&   76.47& 67.88&71.92&   98.00& 89.81&93.73&   93.67& 95.29&94.47
 & 78.12& 60.40& 68.13& 97.60& 59.67&74.06\\ 
         CoCoOp \cite{cocoop}&   80.47& 71.69&75.83&   75.98& 70.43&73.10&   97.96& 93.81&95.84&   95.20& 97.69&96.43
 & 70.49& 73.59& 72.01& 94.87& 71.75&81.71\\
 ProDA \cite{proda}& 81.56& 72.30& 76.65& 75.40& 70.23& 72.72& 98.27& 93.23& 95.68& 95.43& 97.83& 96.62& 74.70& 71.20& 72.91& 97.70& 68.68&80.66\\ 
         KgCoOp \cite{kgcoop}&   80.73& 73.60&77.00&   75.83& 69.96&72.78&   97.72& 94.39&96.03&   94.65& 97.76&96.18 & 71.76& 75.04& 73.36& 95.00& 74.73&83.65\\
 LFA \cite{lfa}& 83.62& 74.56& 78.83& 76.89& 69.36& 72.93& \textbf{98.41}& 93.93& 96.13& 95.13& 96.23& 95.68& 76.32& 74.88& 75.59& 97.34& 75.44&85.00\\
         MaPLe \cite{maple}&   82.28& 75.14&78.55&   76.66& 70.54&73.47&   97.74& 94.36&96.02&   95.43& 97.76&96.58
 & 72.94& 74.00& 73.47& 95.92& 72.46&82.56\\ 
MetaPrompt \cite{metaprompt}& 83.38& 76.09& 79.57& \textbf{77.39}& 71.06& 74.09& 98.28& 94.58& 96.39& 95.71& 96.98& 96.34& 75.43& 74.43& 74.93& 97.53& 74.54&85.50\\
 TCP \cite{tcp}& \textbf{84.13}& 75.36& 79.51& 77.27& 69.87& 73.38& 98.23& 94.67& 96.42& 94.67& 97.20&95.92 & \textbf{80.80}& 74.13& \textbf{77.32}& 97.73& 75.57&85.23\\
MMA \cite{mma}& 83.20& 76.80& 79.87& 77.31& 71.00& 74.02& 98.40& 94.00& 96.15& 95.40& \textbf{98.07}&96.72
 & 78.50& 73.10& 75.70& \textbf{97.77}& 75.93&85.48\\

 \midrule
 DRA (Ours)& 83.06& \textbf{77.75}&\textbf{80.32}& 77.26&\textbf{71.21} & \textbf{74.11}& 98.21& \textbf{94.83}&\textbf{96.49}& \textbf{95.82}& 97.91& \textbf{96.85} & 76.48& \textbf{75.36}& 75.91& 97.25& 76.62&\textbf{85.71}\\
 \midrule
 \midrule
         \multirow{2}{*}{Methods} &  \multicolumn{3}{c|}{Food101}&  \multicolumn{3}{c|}{FGCVAircraft
}&  \multicolumn{3}{c|}{SUN397}&  \multicolumn{3}{c|}{DTD} & \multicolumn{3}{c|}{EuroSAT}& \multicolumn{3}{c}{UCF101}\\ 
          &Base & New & HM& Base & New & HM
& Base & New & HM&Base & New & HM& Base & New & HM& Base & New &HM
\\
          \midrule
         CLIP \cite{clip}
&  90.10& 91.22&90.66&  27.19& 36.29&31.09
&  69.36& 75.35&72.23&  53.24& 59.90&56.37& 56.48& 64.05& 60.03& 70.53& 77.50&73.85
\\ 
         CoOp \cite{coop}
&    
88.33& 82.26&85.19&   40.44& 22.30&28.75
&   
80.60& 65.89&72.51&   79.44& 41.18&54.24& 92.19& 54.74& 68.69& 84.69& 56.05&67.46
\\ 
         CoCoOp \cite{cocoop}
&   
90.70& 91.29&90.99&   33.41& 23.71&27.74
&   79.74& 76.86&78.27&   77.01& 56.00&64.85& 87.49& 60.04& 71.21& 82.33& 73.45&77.64
\\
 ProDA \cite{proda}& 90.30& 88.57& 89.43& 36.90& 34.13& 35.46& 78.67& 76.93& 77.79& 80.67& 56.48& 66.44& 83.90& 66.00& 73.88& 85.23& 71.97&78.04\\ 
KgCoOp \cite{kgcoop}
& 90.50& 91.70&91.09&   36.21& 33.55&34.83&   
80.29& 76.53&78.36&   77.55&54.99&64.35& 85.64& 64.34& 73.48& 82.89& 76.67&79.65\\

 LFA \cite{lfa}& 90.52&91.48 &91.00 &41.48 &32.29 & 36.31& 82.13& 77.20& 79.59& 81.29&60.63 & 69.46&93.40 &71.24 &80.83 & 86.97&77.48 &81.95\\
MaPLe \cite{maple}
&  90.71& \textbf{92.05}&\textbf{91.38}&   37.44& 35.61&36.50
&   80.82& 78.70&79.75&   80.36& 59.18&68.16& \textbf{94.07}& 73.23& 82.35& 83.00& 78.66&80.77
\\
MetaPrompt \cite{metaprompt}& \textbf{90.76}& 91.77& 91.26& 39.38 &37.59 & 38.46& 82.10& 79.01&80.53 &82.52 &60.10 &69.55 &93.37 &78.03 &86.04 &87.61 &81.18 &84.21\\
 TCP \cite{tcp}
&  90.57& 91.37& 90.97& \textbf{41.97} & 34.43 & 37.83& 
\textbf{82.63}& 78.20 & 80.35 &82.77 &58.07 &68.25 & 91.63& 74.73& 82.32& 87.13 & \textbf{80.77} &\textbf{83.83}\\
MMA \cite{mma}
& 90.13& 91.30& 90.71& 40.57& 36.33& 38.33
& 82.27& 78.57& 80.38& \textbf{83.20}& 65.63&73.38& 85.46& 82.34& 83.87& 86.23& 80.03&82.20
\\
 \midrule
 DRA (Ours)&  
90.53&91.78&91.15& 39.42 & \textbf{38.59} & \textbf{39.00} &82.50&\textbf{79.29}&\textbf{80.86}&83.18&\textbf{67.07} &\textbf{74.26} &85.83 &\textbf{82.72} &\textbf{84.25} &\textbf{87.21} &79.85&83.37\\
\bottomrule
    \end{tabular}
\end{adjustbox}
\end{table*}

\subsection{Experimental Settings}
\label{sec:experimentsetting}
\subsubsection{Datasets}
To verify the effectiveness of our proposed method, following the benchmark adopted by previous works \cite{coop,kgcoop,mma}, we conducted experiments on Base-New Generalization, Cross-datasets Evaluation and Domain Generalization. In base-new generalization and cross-datasets benchmarks, we used a total of 11 datasets of different types, including object recognition datasets: ImageNet \cite{imagenet} and Caltech101 \cite{caltech101}; five fine-grained visual classification datasets: Oxford Pets \cite{oxfordpet}, Stanford Cars \cite{oxfordpet}, Flowers 102 \cite{flower}, Food 101 \cite{food}, and FGVC Aircraft \cite{fgvcaircraft}; scene understanding dataset SUN397 \cite{sun397}; material dataset DTD \cite{dtd}; satellite image recognition dataset EuroSAT \cite{eurosat}; and action classification dataset UCF101 \cite{ucf101}. These datasets can evaluate the generalization ability of the model comprehensively.

\subsubsection{Implementation Details}

Following previous works \cite{coop,cocoop,mma}, we adopt the few-shot learning setting with only 16 shots per class when fine-tuning in all benchmarks. We adopt the pre-trained ViT-B/16 by CLIP as the backbone network for all experiments. We set the batch size to 16, except for ImageNet, which is 128. We adopt the SGD optimizer with a cosine annealing strategy and set the initial learning rate to 0.0015. We use "a photo of a [\texttt{class}]" as the word embedding consistent with zero-shot learning. For base-new generalization benchmark, we fine-tune 5 epochs for all datasets, and the starting number of layers $h$ added to DRA is set to 5. For cross-dataset evaluation and domain generalization benchmarks, we fine-tune Imagnet for one epoch, and the starting number of layers added to DRA is set to 7. For our proposed DRA, the intermediate dimension, i.e., the rank of the feature $r$ is set to 32, and the scaling factor $s$ is set to 0.001. The proposed DRA is applied to each attention module in each transformer. We set the number of token importance groups $K$ to 4 and the rank ratio is set to [1.0, 0.8, 0.6, 0.4]. The loss coefficients $\lambda_T$ and $\lambda_V$ are both set to 1. The final performance is averaged over three random seeds. All experiments are conducted on a RTX 4090.

\subsection{Comparison with SOTA Methods}
\label{sec:sota}
\subsubsection{Base-New Generalization}
To verify the generalization ability of our proposed DRA, Base-New Generalization classifies the dataset into two disjoint subsets, base classes and new classes. The base classes are used for fine-tuning, and the fine-tuned model is directly tested on the new classes. Table \ref{tab:main} reports the comparison results with existing methods, where the base, new, and HM represent the base class accuracy, new class accuracy, and the harmonic mean of the two, respectively. Experimental results show that the new classes and HM of our proposed method outperform the SOTA methods in average performance on 11 datasets. Among the 11 datasets, DRA achieves better performance in 7/11 new classes and higher harmonic mean accuracy in 8/11 datasets compared with the SOTA methods. Specifically, compared with CLIP, the proposed method achieves higher accuracy on all base classes and 10/11 new classes. Existing prompt-based methods such as CoOp, CoCoOp, KgCoOp, and TCP generally achieve a larger improvement in base classes compared with CLIP but perform poorly in new classes. ProDA and LFA focus on fine-tuning VLMs on the distribution of text and image features, but they also perform poorly on new classes. This is because the existing methods overfit the base class training data and have difficulty distinguishing discriminative information from noise information in the tokens. Compared with MaPLe, which uses learnable prompts for multimodal interaction, DRA improves the accuracy of base classes and new classes by an average of 0.78\% and 2.61\%, respectively. Compared with MMA, which also uses the Adapter architecture, DRA focuses on improving the generalization ability of new classes by introducing the token importance perspective and uses a channel response mechanism to alleviate the potential impact on the performance of the base class. In the end, DRA improved the accuracy of new classes by 0.95\%, while the accuracy of the base class was only a small difference of 0.14\%. These results show that DRA can effectively improve the generalization ability of the model for new classes.

\begin{table*}[ht]
    \centering
    \caption{Comparison Results in Cross-dataset Evaluation.}
    \label{tab:cross}
    \begin{tabular}{rc|ccccccccccc}
    \toprule
         \multirow{2}{*}{Method}& Source & \multicolumn{11}{c}{Target}\\
         \cmidrule{2-13}
         &  ImageNet&  Caltech&  Pets&  Cars&  Flowers&  Food&  Aircraft&  SUN&  DTD& EuroSAT& UCF&Average\\
         \midrule
         CoOp \cite{coop}&  \textbf{71.51}&  93.70&  89.14&  64.51&  68.71&  85.30&  18.47&  64.15&  41.92& 46.39& 66.55&63.88
\\
         CoCoOp \cite{cocoop}&  71.02&  \textbf{94.43}&  90.14&  65.32&  71.88&  86.06&  22.94&  67.36&  45.73& 45.37& 68.21&65.74\\
 MaPLe \cite{maple}& 70.72& 93.53& 90.49& 65.57& 72.23& 86.20& 24.74& 67.01& 46.49& 48.06& 68.69&66.30
\\
 TCP \cite{tcp}& 71.40& 93.97& \textbf{91.25}& 64.69& 71.21& \textbf{86.69}& 23.45& 67.15& 44.35& \textbf{51.45}& 68.73&66.29\\
 MMA \cite{mma}& 71.00& 93.80& 90.30& 66.13& 72.07& 86.12&\textbf{25.33}& 68.17& \textbf{46.57}& 49.24& 68.32&66.61
\\
 \midrule
DRA (Ours)& 71.49& 94.23& 90.53& \textbf{66.65}& \textbf{72.36}& 86.44& 25.28& \textbf{68.29}& 45.94& 47.63& \textbf{69.71}&\textbf{66.71}\\
 \bottomrule
    \end{tabular}
\end{table*}

\subsubsection{Cross-Dataset Evaluation}

We validated our proposed DRA on the Cross-Datasets Evaluation benchmark, and the results are shown in Table \ref{tab:cross}. The model is first fine-tuned on ImageNet and directly tested on 10 target datasets. Our DRA achieves the highest average precision compared with the state-of-the-art methods. Specifically, DRA outperforms CoCoOp over 9/10 datasets. Compared to MaPLe, the proposed DRA leads in 8/10 datasets, while it outperforms TCP and MMA in 7/10 datasets. In addition, DRA achieves competitive results on ImageNet, outperforming other methods except CoOp. These results demonstrate the advantage of our method in zero-shot transfer capability in various downstream tasks.

\begin{table}[ht]
    \centering
        \caption{Comparison Results in Domain Generalization.}
    \label{tab:ood}
    \setlength{\tabcolsep}{4pt}
    \begin{tabular}{rc|cccc|c}
    \toprule
         Method&  ImgNet&  -V2&  -S&  -A& -R &Avg\\
         \midrule
         CLIP \cite{clip}&  66.73&  60.83&  46.15&  47.77& 73.96 &57.18\\
         CoOp \cite{coop}&  \textbf{71.51}&  64.20&  47.99&  49.71& 75.21 &59.28\\
         CoCoOp \cite{cocoop}&  71.02&  64.07&  48.75&  50.63& 76.18 &59.91\\
 KgCoOp \cite{kgcoop}& 71.20& 64.10& 48.97& 50.69&76.70 &60.12\\
         MaPLe \cite{maple}&  70.72&  64.07&  49.15&  50.90& 76.98 &60.28\\
         MMA \cite{mma}&  71.00&  64.33&  49.13&  \textbf{51.12}& 77.32 &60.48\\
         \midrule
         DRA (Ours)&  71.49&  \textbf{64.64}&  \textbf{49.60}&  50.75& \textbf{77.58} &\textbf{60.64}\\
         \bottomrule
    \end{tabular}

\end{table}

\subsubsection{Domain Generalization}
Following previous works \cite{coop,cocoop}, we fine-tune the model on ImageNet and directly test on 4 out-of-distribution datasets to verify the domain generalization performance of the method, and the results are shown in Table \ref{tab:ood}. The proposed DRA achieves the highest average precision and exceeds the existing SOTA performance in 3/4 ImageNet variant datasets. These results demonstrate the robustness of our proposed method in the face of domain shift.

\begin{table}[ht]
    \centering
    \caption{Results of the Component Ablation Experiments.}
    \label{tab:ablation}
    \begin{tabular}{lllccc}
    \toprule
         \multirow{2}{*}{Method} & \multirow{2}{*}{CR}&\multirow{2}{*}{$\mathcal{L}_{reg}$}&  \multicolumn{3}{c}{Accuracy (\%)}\\
         \cmidrule{4-6}& &&  Base&  New& H Mean\\
         \midrule
         Baseline & -&-&  82.79&  76.29& 79.41\\
         DRA-T & \checkmark&\checkmark&  82.85&  76.68& 79.65\\
         DRA-V & \checkmark&\checkmark&  \textbf{83.24}&  76.88& 79.91\\
         \midrule
 DRA-T\&V & -& -& 82.92& 76.81&79.74\\
         DRA-T\&V & \checkmark&-&  83.00&  77.42& 80.12\\
 DRA-T\&V & -& \checkmark& 83.01& 77.05&79.92\\
         \midrule
         DRA-T\&V& \checkmark&\checkmark&  83.06&  \textbf{77.75}& \textbf{80.32}\\
         \bottomrule
    \end{tabular}
\end{table}
\subsection{Ablation Studies}
\label{sec:ablation}

\begin{figure}
    \centering
    \includegraphics[width=1\linewidth]{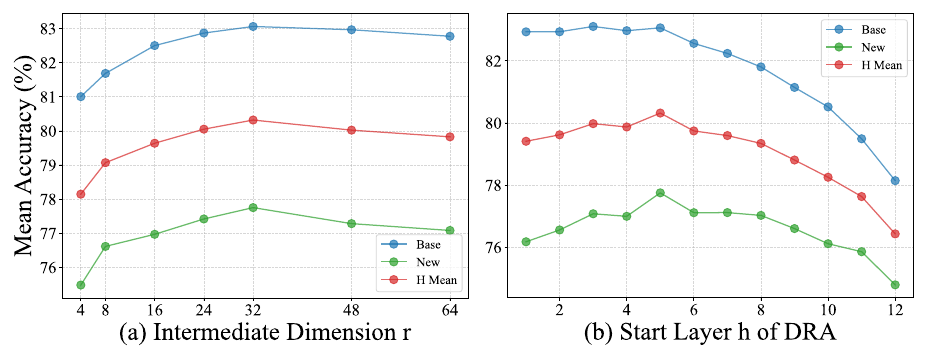}
    \caption{Results of ablation experiments on the impact of intermediate dimensions $r$ (a) and starting layers $h$ (b).}
    \label{fig:hyper}
\end{figure}
\subsubsection{Component Ablation}
To verify the effectiveness of our proposed DRA, we validated the effectiveness of our proposed module. The results are shown in Table \ref{tab:ablation}. First, we added the full DRA to the text and image branches respectively. Compared with the standard Adapter architecture, the accuracy of both the base class and the new class was improved after adding DRA. DRA-V achieved a greater improvement than the Baseline, especially the base class performance, which indicates that visual features may contain more redundant information in the base class that can be effectively reduced by DRA. In addition, when CR and regularization loss are removed, and DRA is applied to both the image and text branches, HM has a 0.33\% improvement over the Baseline, which proves the effectiveness of the basic dynamic rank assignment. After adding CR, HM has shown a significant improvement because CR allows the dynamic selection of important channels to be retained according to different samples during rank adaptation. If only the regularization loss is retained without using CR, the model also performs better than the basic DRA-T\&V on the New class and HM (79.92\%), indicating that the loss helps stabilize training and promote generalization. When all the proposed modules work together, our full DRA method achieves the best performance in both New class accuracy and harmonic mean HM.

\begin{table}[ht]
    \caption{Results of different token groups and rank ratios}
    \label{tab:group}
    \centering
    \begin{tabular}{cl|ccc}
    \toprule
        $K$ & $[e_1,\dots,e_K]$ & Base  &New  & HM\\
         \midrule
        2 & [1.0, 0.5] & \textbf{83.18} & 77.41 & 80.19\\
        2 & [1.0, 0.25] & 82.59 & 77.25 & 79.83\\
        \midrule
        3 & [1.0, 0.7, 0.4] &83.08  & 77.56  & 80.23\\
        3 & [1.0, 0.6, 0.2] & 82.06 & 77.39 & 79.66\\
        \midrule
        4 & [1.0, 0.8, 0.6, 0.4] & 83.06 & \textbf{77.75} & \textbf{80.32}\\
        4 & [1.0, 0.7, 0.4, 0.1] & 82.17 & 77.01 & 79.51\\
        \midrule
        5 &[1.0, 0.8, 0.6, 0.4, 0.2] & 82.69 & 77.51 & 80.02 \\
         \bottomrule
    \end{tabular}
\end{table}

\subsubsection{Impact of Number of Tokens Groups and Rank Ratio}
We further investigated the impact of the number of groups $K$ in token importance grouping and the corresponding rank ratios $[e_1, ..., e_K]$ in rank adaptation, and the results are detailed in Table \ref{tab:group}. From the results, we can observe that increasing the number of groups from 2 to 4 can slightly improve the performance, particularly regarding the harmonic mean (HM) and new class accuracy. This suggests that a more fine-grained differentiation of tokens based on their importance can lead to a more efficient adaptation process.  When the number of groups is the same, higher rank ratios perform significantly better than lower rank ratios. These results suggest that although compressing less informative tokens is beneficial, over-compression may lead to the loss of potentially valuable contextual information and hinder model generalization. Therefore, choosing an appropriate number of groups combined with a moderate rank ratio can achieve the best balance to maximize overall performance.

\subsubsection{Impact of Intermediate Dimension}
We further studied the impact of the choice of the highest rank $r$ (the intermediate dimension) in the DRA module on the model performance, and the results are shown in Figure \ref{fig:hyper}(a). The experimental results show that as the value of $r$ increases, the model is able to capture richer adaptive information, and the accuracy of both the base class and the new class is steadily improved. This trend continues until $r=32$ when the model reaches a peak in the harmonic mean accuracy. However, when the intermediate dimension continues to increase, we observe that the performance begins to decline, especially the new class accuracy. This indicates that too high an intermediate dimension may increase the risk of overfitting the base class training data and capture some features or noise that are not conducive to generalization to new classes, thereby damaging the performance of new classes. 
Therefore, $r=32$ was determined to be the optimal intermediate dimension setting.

\subsubsection{Impact of Starting Layer}
The results of the impact of the start layer $h$ of the DRA module are shown in Figure \ref{fig:hyper}(b). When DRA is only applied to the last few layers of the network, the performance improvement is relatively limited. As we gradually extend the application scope of DRA forward to deeper layers, the accuracy of both base and new classes shows an upward trend. When the $h=5$, both the base and new classes achieve high performance. However, if DRA is continued to be applied to earlier layers, 
the new class accuracy begins to decline. This suggests that the premature introduction of parameterized fine-tuning modules at the bottom of the network may begin to interfere with the general visual representation in the pre-trained model that is crucial for the generalization of new classes, thereby reducing the performance of new classes.

\subsubsection{Analysis of Computational Efficiency}
We further evaluate the computational efficiency performance of the proposed DRA and other methods in terms of training epoch, training FPS per epoch and inference FPS in Table \ref{tab:efficiency}. In terms of number of parameters, DRA introduces 0.717M trainable parameters, which is significantly less than MaPLe (3.55M), but more than CoOp, KgCoOp (all 0.002M), and CoCoOp (0.354M). Despite the moderate number of parameters, DRA converges in only 5 fine-tuning epochs, which is much less than CoOp, KgCoOp and CoCoOp. In terms of training FPS per epoch, DRA (138.73 FPS) is comparable to MaPLe (138.26 FPS), significantly faster than CoCoOp (16.80 FPS), but slower than CoOp (250.12 FPS) and KgCoOp (266.29 FPS). In terms of inference speed, DRA (642.90 FPS) is faster than CoCoOp (166.64 FPS) but slightly slower than CoOp (888.46 FPS), KgCoOp (957.70 FPS), and MaPLe (920.20 FPS). This shows that DRA obtains an excellent balance between efficiency and performance.
\begin{table}[t]
    \caption{Comparison Results of computational efficiency.}
    \label{tab:efficiency}
    \centering
    \begin{adjustbox}{width=\linewidth}
    \begin{tabular}{rccccc}
    \toprule
         \multirow{2}{*}{Method}& \multirow{2}{*}{Param.}  &\multirow{2}{*}{Epoch}  & \multicolumn{2}{c}{FPS} & \multirow{2}{*}{HM (\%)} \\
    \cmidrule{4-5}&   && Train & Inference &  \\
         \midrule
        CoOp \cite{coop} & 0.002M &50& 250.12&  888.46& 71.66\\
         CoCoOp \cite{cocoop}&  0.354M &10&  16.80&  166.64& 75.83\\
         KgCoOp \cite{kgcoop}&  0.002M &100&  266.29&  957.70& 77.00\\
         MaPLe \cite{maple}&  3.555M &5&  138.26&  920.20& 78.55\\
         \midrule
         DRA (Ours) &  0.717M &5&  138.73&  642.90& 80.32\\
         \bottomrule
    \end{tabular}
    \end{adjustbox}
\end{table}
\begin{figure}
    \centering
    \includegraphics[width=1\linewidth]{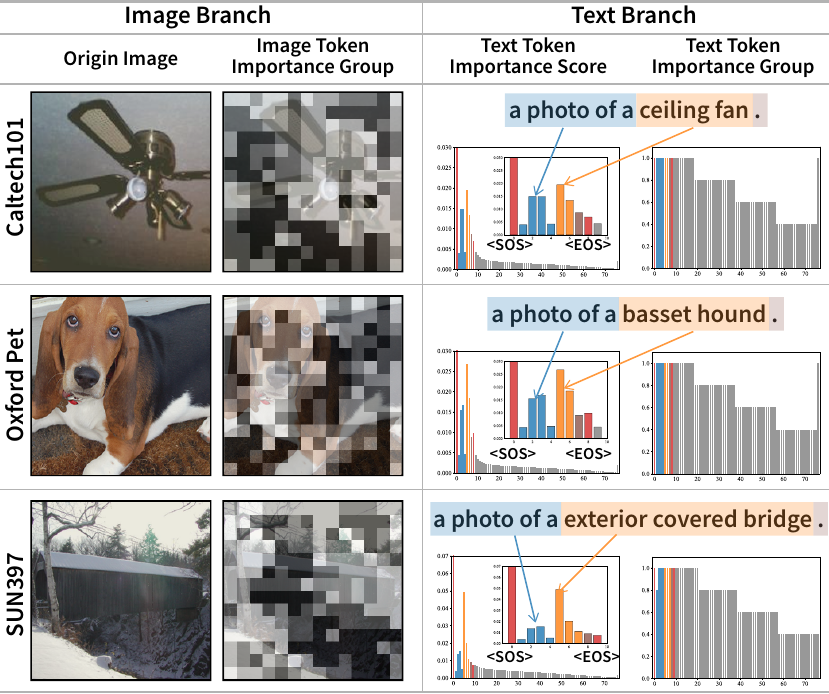}
    \caption{Visualization of the token importance group and feature rank allocation for image and text tokens.}
    \label{fig:visual}
\end{figure}

\subsection{Visualization Analysis}
\label{sec:visual}
We visualize the distribution of token importance and feature rank of the tokens in image and text modalities, as shown in Figure \ref{fig:visual}. In the image branch, lighter colors indicate that the token is assigned a higher feature rank. Benefiting from our proposed token importance grouping, the foreground and discriminative regions of the sample are assigned higher ranks, while the background is assigned a lower rank, thereby reducing overfitting. In the text branch, since the token score of the start token \texttt{<SOS>} is too high (usually greater than 0.7), we crop the histogram for easier viewing. We can observe that lower ranks are assigned to classification-irrelevant prepositions such as ‘a’ and blank word vectors during text tokenization. Compared with the image branch, the effective information of the text branch is more concentrated at the front of the text sequence. The above results verify the effectiveness of our method and help further analyze our proposed DRA.

\section{Conclusion}
\label{sec:conclusion}
In this paper, we proposed Dynamic Rank Adaptation (DRA), a novel Adapter-based fine-tuning approach to address the challenge of preserving and enhancing the generalization ability of pre-trained Vision-Language Models (VLMs) to unseen new classes. The core of DRA is to dynamically allocate higher effective ranks to more important tokens while compressing less significant tokens to preserve the general representations vital for novel concept recognition. DRA introduces the token importance grouping by sequence attention and divides tokens with different importance. Furthermore, a channel response mechanism is employed to prioritize preserving the most informative feature channels during rank adaptation, particularly mitigating potential information loss when reducing rank. An L1 regularization term was also incorporated to ensure training stability. Extensive experiments demonstrated the superiority of DRA, especially in new classes.

\bibliographystyle{ACM-Reference-Format}
\bibliography{DRA.bib}
\end{document}